\documentclass[10pt,twocolumn,letterpaper]{article}

\usepackage{iccv}
\usepackage{times}
\usepackage{epsfig}
\usepackage{graphicx}
\usepackage{amsmath}
\usepackage{amssymb}
\usepackage{booktabs}
\usepackage{tablefootnote}
\usepackage{threeparttable}
\usepackage{enumitem}
\usepackage{times}
\usepackage{stfloats}
\usepackage{amsfonts}
\usepackage{bbm}
\usepackage{bm}
\usepackage{multirow}
\usepackage{makecell}
\usepackage{ulem}
\usepackage{bbding}
\usepackage{subfig}
\usepackage{float}
\usepackage[accsupp]{axessibility}
\usepackage[table]{xcolor}
\usepackage[bottom]{footmisc}
\newcommand{\ignore}[1]{}
\definecolor{cGreen}{RGB}{100,180,100}
\definecolor{cRed}{RGB}{220,50,0}
\definecolor{Klein_Blue}{rgb}{0.0, 0.129, 0.6}
\usepackage[pagebackref,breaklinks,colorlinks]{hyperref}
\hypersetup{
    colorlinks=true,
    linkcolor=magenta,
    urlcolor=magenta,
    citecolor=Klein_Blue
    }
    \usepackage[page]{appendix}
\usepackage[capitalize]{cleveref}
\crefname{section}{Sec.}{Secs.}
\Crefname{section}{Section}{Sections}
\Crefname{table}{Table}{Tables}
\crefname{table}{Tab.}{Tabs.}
\iccvfinalcopy 
\def\iccvPaperID{5473} 

\ificcvfinal\fi

\begin{document}

\title{Exploring Lightweight Hierarchical Vision Transformers  \\ for Efficient Visual Tracking}

\author{
Ben Kang$^{1,*}$ \ \ \ \ Xin Chen$^{1,*}$   \ \ \ \ Dong Wang$^{1,\dagger}$ \ \ \ \ Houwen Peng$^{2,\dagger}$ \ \ \ \ Huchuan Lu$^{1}$ \\
$^1$ School of Information and Communication Engineering, Dalian University of Technology \\ $^2$ Microsoft Research}

%\small
%\vspace{-0.4em} 
%\texttt{\{kangben,chenxin3131\}@mail.dlut.edu.cn} \ \ \ \ \ 
%\texttt{\{wdice\}@dlut.edu.cn} \ \ \ \ \  
%\texttt{\{penghouwen\}@icloud.com} \\ 
%\small
%\vspace{-2.5em} 
%\texttt{\{lhchuan\}@dlut.edu.cn} 

\maketitle
\ificcvfinal\thispagestyle{empty}\fi

\begin{abstract}
    Transformer-based visual trackers have demonstrated significant progress owing to their superior modeling capabilities. However, existing trackers are hampered by low speed, limiting their applicability on devices with limited computational power. To alleviate this problem, we propose HiT, a new family of efficient tracking models that can run at high speed on different devices while retaining high performance. 
    The central idea of HiT is the Bridge Module, which bridges the gap between modern lightweight transformers and the tracking framework. The Bridge Module incorporates the high-level information of deep features into the shallow large-resolution features. In this way, it produces better features for the tracking head. We also propose a novel dual-image position encoding technique that simultaneously encodes the position information of both the search region and template images. The HiT model achieves promising speed with competitive performance. For instance, it runs at 61 frames per second (fps) on the Nvidia Jetson AGX edge device. Furthermore, HiT attains 64.6\% AUC on the LaSOT benchmark, surpassing all previous efficient trackers. Code and models are available at \href{https://github.com/kangben258/HiT}{https://github.com/kangben258/HiT}.
\end{abstract}

\newcommand\blfootnote[1]{% 
\begingroup 
\renewcommand\thefootnote{}\footnote{#1}% 
\addtocounter{footnote}{-1}% 
\endgroup 
}
{
	\noindent \blfootnote{\hspace{-6mm}$^{*}$ Equal contribution.
	
	\hspace{-2.3mm}$^\dagger$ Corresponding authors: Dong Wang (wdice@dlut.edu.cn), Houwen Peng (houwen.peng@microsoft.com).
	}
}

\vspace{-6mm}
\section{Introduction}
\label{sec:intro}
Visual object tracking is a fundamental task in computer vision, which aims to track an arbitrary object given its initial state in a video sequence. In recent years, with the development of deep neural networks~\cite{AlexNet,ResNet,googlenet,2017Attention}, tracking has made significant progress. In particular, the utilization of transformers~\cite{2017Attention} has played a pivotal role in the development of several high-performance trackers~\cite{TransT,Stark,wang2021transformer,yu2021high,mixformer,ostrack,SeqTrack}. Unfortunately, much of the research~\cite{SiameseRPN,DiMP,TransT} has concentrated solely on achieving high performance without considering tracking speed. While these trackers may achieve real-time speed on powerful GPUs, they lack competitiveness and advantages on resource-limited devices.
\begin{figure}[!t]
\begin{center}
\includegraphics[width=1.0\linewidth]{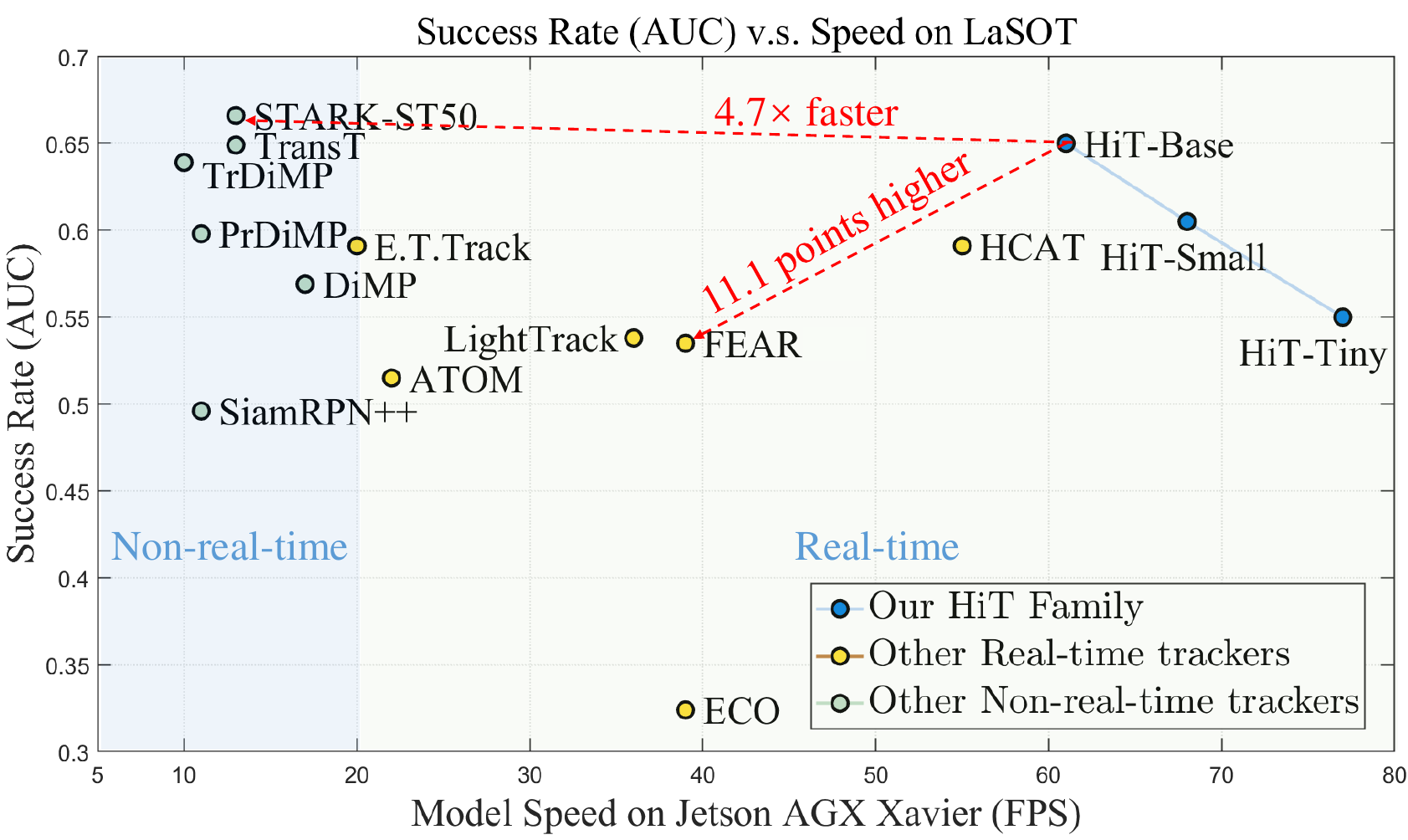}
\end{center}
   \vspace{-3mm}
   \caption{Comparison of our HiT with other trackers on LaSOT in terms of speed (horizontal axis) on the edge AI platform of Nvidia Jetson AGX Xavier and success rate (AUC) (vertical axis). Following the VOT real-time setting~\cite{vot2020}, we set the real-time line at 20 \emph{fps}. Our HiT achieves the best real-time result, surpassing other efficient trackers.}
      \vspace{-3mm}
\label{fig:compare on last}
\end{figure}
For instance, TransT~\cite{TransT}, which is a high-performance tracker, only achieves a speed of 5 frames per second (\emph{fps}) on the Intel Core i9-9900K CPU and 13 \emph{fps} on the Nvidia Jetson AGX. Consequently, a high-performance tracker with fast speed is critical. 

The one-stream structure has gained popularity in tracking applications~\cite{ostrack,simtrack,sbt,mixformer}. This structure performs feature extraction and feature fusion jointly, leveraging the capabilities of the backbone network~\cite{ViT} that has been pre-trained for image classification. In our work, we also adopt the one-stream architecture, leveraging a pre-trained lightweight transformer backbone network. However, there exists a substantial gap between the tracking field and the image classification field. In the image classification field, lightweight networks~\cite{graham2021levit,mehta2021mobilevit,wu2022tinyvit} frequently incorporate a hierarchical architecture with high-stride downsampling to decrease computational expenses. However, large-stride downsampling often leads to a loss of critical information, which is crucial for accurate tracking. This naturally raises the question of how to reconcile the need for detailed information in tracking with the large-stride downsampling in the hierarchical backbone network. 

To tackle this problem, we introduce the Bridge Module, which integrates features from different levels of the hierarchical backbone. The Bridge Module fuses deep semantic information with shallow detail information, mitigating the information loss resulting from large-stride downsampling.  By combining the proposed Bridge Module with lightweight hierarchical backbone LeViT~\cite{graham2021levit}, we develop HiT, a new family of efficient tracking models.
Moreover, we proposed a novel relative position encoding technique, called dual-image position encoding, to improve the position information. This method encodes the position information of the template and search region jointly, enhancing the interaction between them. 

Our extensive experiments validate the effectiveness and efficiency of HiT. Specifically, compared to the high-speed tracker FEAR~\cite{lin2017feature}, HiT-Base achieves an 11.1\% higher AUC score on the LaSOT benchmark while being 1.6 times faster than FEAR on Nvidia Jetson AGX Xavier. In comparison to the high-performance tracker STARK-ST50~\cite{Stark}, HiT-Base delivers similar performance while being 4.7 times faster on AGX, representing a significant improvement over previous real-time trackers. Our main contributions are summarized as follows:
\begin{itemize}[leftmargin=0.468cm]
\vspace{-1mm}
    \item{We propose the Bridge Module, which incorporates the high-level information of deep features into the shallow large-resolution features, thereby mitigating the information loss caused by the large-stride downsampling. This approach enables the use of large-stride downsampling hierarchical backbones for tracking purposes.}
    \item{To improve position accuracy, we introduce a dual-image position encoding approach that jointly encodes position information from both the template and the search region.}
    \item{Building upon these components, we introduce HiT, a family of efficient tracking models. HiT exhibits promising performance while maintaining exceptionally fast processing speeds. Empirical evaluations demonstrate that HiT outperforms state-of-the-art efficient tracking algorithms.}
\end{itemize} 

\vspace{-3mm}
\section{Related Work}
\label{sec:related work}

\textit{Visual Tracking.} Siamese-based methods~\cite{SiameseFC,SINT,SiameseRPN,SiamMask,SiamRPNplusplus,SiamFC++,SiamCAR,SiamBAN,Deeper-wider-SiamRPN} are popular in tracking. The Siamese-based framework typically employs two backbone networks with shared parameters to extract the features of the template and the search region images, uses a correlation-based network for feature interaction, and finally uses head networks for prediction. TransT~\cite{TransT}, TMT~\cite{wang2021transformer}, and their follow-up works~\cite{Stark,liu2021swin,ToMP,CSWinTT,AiATrack} further improve tracking performance by introducing the transformer~\cite{2017Attention} for the feature interaction.
Recently, a one-stream framework establishes new state-of-the-art performance in tracking, such as MixFormer~\cite{mixformer}, SBT~\cite{sbt}, SimTrack~\cite{simtrack}, and OSTrack~\cite{ostrack}. The one-stream framework jointly performs feature extraction and feature fusion with the backbone network. This framework is simple yet effective by leveraging the capabilities of the backbone network that has been pre-trained for image classification. However, these methods are developed for powerful GPUs, and their speeds on edge devices are slow, limiting their applicability. In this work, we also adopt the one-stream framework and we focus on making this framework more efficient.

\textit{Efficient Tracking Network.} Practical applications require efficient trackers that can achieve high performance and fast speed on edge devices. Early methods ECO~\cite{danelljan2017eco} and ATOM~\cite{ATOM} achieve real-time speed on edge devices, but the performance is inferior compared with current state-of-the-art trackers. Recently, some efficient trackers have been developed. LightTrack~\cite{yan2021lighttrack} uses NAS to search networks, which entails a low computational amount and relatively high performance. FEAR~\cite{borsuk2022fear} obtains a family of efficient and accurate trackers by employing a dual-template representation and a pixel-wise fusion block. However, there is still a large performance gap between these efficient trackers and the popular high-performance trackers~\cite{TransT,Stark}. In this work, the proposed HiT not only runs at high speed on edge devices but also achieves competitive results compared with high-performance trackers.
For example, compared with TransT~\cite{TransT}, our method performs only 0.3\% lower (in AUC) on LaSOT but 4.7 times faster on AGX.

\textit{Vision Transformer.} ViT~\cite{ViT} introduces the transformer to image classification and has achieved impressive performance. After that, a large number of vision transformer networks~\cite{touvron2021training,yuan2021tokens,wu2021cvt,wang2021pyramid,liu2021swin} are developed. Transformers are popular for their superior modeling capabilities but are limited in speed. Therefore, many lightweight vision transformers~\cite{mehta2021mobilevit,graham2021levit,wu2022tinyvit} have emerged, greatly accelerating the speed of transformer-based networks. Different from the classic vision transformer, these lightweight transformers employ a hierarchical architecture with high-stride downsampling to decrease computational expenses.
In this work, we focus on leveraging the lightweight hierarchical vision transformer with the one-stream tracking framework. By default, we employ LeViT~\cite{graham2021levit} as the backbone network. However, our method has the following fundamental differences from LeViT. (1) The overall architectures are different. LeViT makes predictions on the final heavily-downsampled features. Our HiT employs a Bridge Module to fuse features of different stages, and the predictions are made on the fused large-resolution features. We also modify the transformer module so that it can handle the search region and template simultaneously.
(2) The tasks are different. LeViT is designed for image classification, focusing more on high-level semantic information. Our framework is for tracking, where detailed information is also crucial. 
(3) The position encodings are different. LeViT encodes the position information for a single image. We develop the dual-image position encoding to encode the position information of the template and search region jointly to enhance the level of detail.

\begin{figure*}[t]
\begin{center}
\includegraphics[width=1\linewidth]{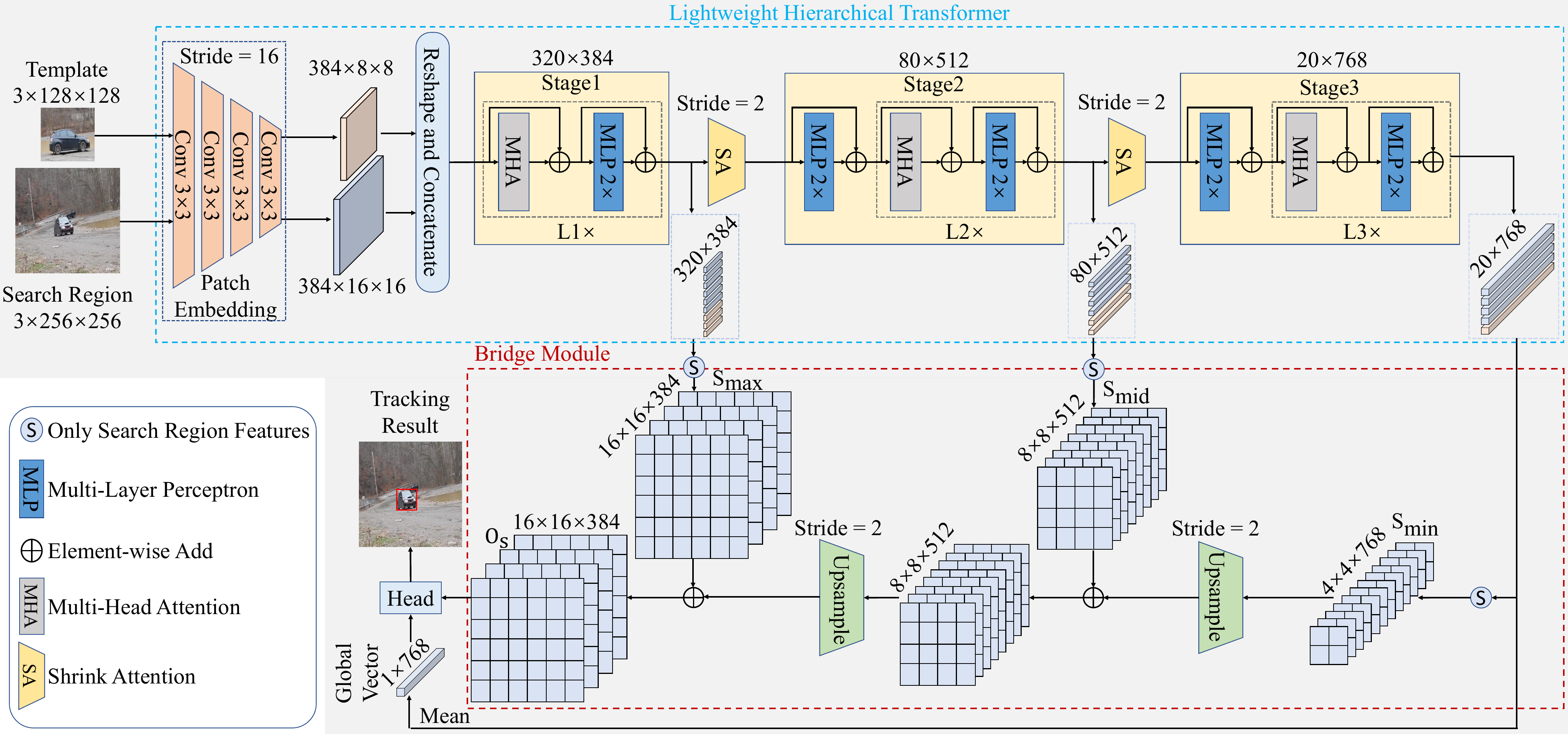}
\end{center}
    \vspace{-3mm}
   \caption{Architecture of the proposed HiT framework. The HiT framework contains three components: a lightweight hierarchical vision transformer for feature extraction and fusion, a Bridge Module that combines multi-stage features, and a prediction head. }
\vspace{-3mm}
\label{fig:framework}
\end{figure*}

\section{Method}
\label{sec:Method}
This section presents the HiT method in detail.
First, we briefly overview our HiT framework. Then, we depict the model architecture, including the lightweight hierarchical vision transformer with our dual-image position encoding, the proposed Bridge Module, and the head network. Finally, we introduce the training and inference pipelines.

\subsection{Overview}
As shown in Figure~\ref{fig:framework}, HiT is a one-stream tracking framework consisting of three components: the lightweight hierarchical transformer, the proposed Bridge Module, and the head network. The image pair (including the search region and template images) are fed into the lightweight hierarchical transformer for feature extraction and feature fusion. The core modules of the hierarchical vision transformer are the Multi-Head Attention (MHA), the Shrink Attention (SA), and the dual-image position encoding. MHA extracts and fuses the features of the search and template images, SA downsamples the features, and dual-image position encoding encodes the position information of the search and template images jointly.
From each stage of the hierarchical transformer, we obtain a sequence of features with different resolutions. 
From the last stage of the hierarchical transformer, we obtain a global vector by averaging the final output features.
The feature sequence is input to the Bridge Module, in which features are fused to obtain enhanced features. 
Finally, the global vector and the enhanced features are input into the prediction head to obtain the tracking result. 

\subsection{Lightweight Hierarchical Vision Transformer}
\textit{Hierarchical Backbone.} We use LeViT~\cite{graham2021levit}, a lightweight hierarchical vision transformer as the backbone of HiT. 
We adapt it into our tracking framework. Specifically, the input of the transformer is the template image $\mathbf{Z} \in {\mathbb{R}}^{3 \times {H_{z}} \times {W_{z}}}$ and the search region image $\mathbf{X} \in {\mathbb{R}}^{3 \times {H_{x}} \times {W_{x}}}$. 
First, downsample the image pair by a factor of 16 through patch embedding to get $\mathbf{Z_{p}} \in {\mathbb{R}}^{C \times {\frac{H_{z}}{16}} \times {\frac{W_{z}}{16}}}$ and $\mathbf{X_{p}} \in {\mathbb{R}}^{C \times {\frac{H_{x}}{16}} \times {\frac{W_{x}}{16}}}$. 
Then $\mathbf{Z_{p}}$ and $\mathbf{X_{p}}$ are flattened and concatenated in the spatial dimension and then fed into the following hierarchical transformer. 
The hierarchical transformer consists of three stages. The $i$$-$$th$ stage has $Li$ blocks ($L1$$=$$L2$$=$$L3$$=$$4$, by defaut). Each block consists of a Multi-Head Attention and an MLP in the residual form. Shrink Attention modules are used to connect each stage, and it downsamples the features by a factor of 4 in the spatial dimension. For the output features of each stage, we get the partial features corresponding to the search image. %which will be fed into the Bridge Module.
For the final stage, we also average its output features and get a global vector $\mathbf{G}$. After the transformer backbone, we obtain a global vector $\mathbf{G} \in {\mathbb{R}}^{1 \times C_{min}}$ and a feature sequence with three feature maps of different size: $\mathbf{S_{max}} \in  {\mathbb{R}}^{{H_{max}} \times W_{max} \times C_{max}}$, $\mathbf{S_{mid}} \in {\mathbb{R}}^{{H_{mid}} \times W_{mid} \times C_{mid}}$, $\mathbf{S_{min}} \in {\mathbb{R}}^{{H_{min}} \times W_{min} \times C_{min}}$, where ${C_{max}}=384$, ${C_{mid}}=512$, ${C_{min}}=768$.

\begin{figure}[t]
	\centering
	\subfloat[Multi-Head Attention (MHA)]{\label{fig:MHA}\includegraphics[width=0.6\linewidth]{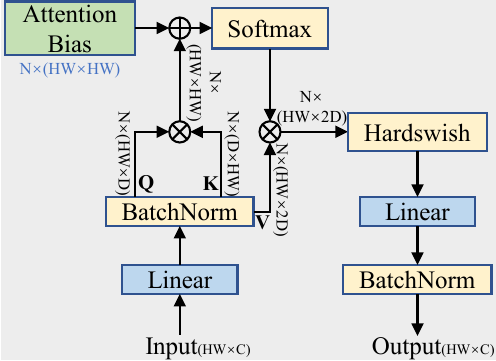}}\\
 	\vspace{-3mm}
	\subfloat[Shrink Attention (SA)]{\label{fig:Shrink}\includegraphics[width=0.8\linewidth]{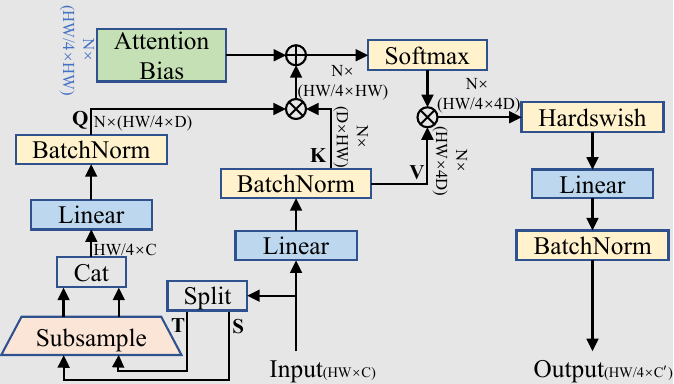}}\\	
 	\vspace{-3mm}
	\caption{Detailed architectures of MHA and SA.}
	\vspace{-5mm}
\end{figure}

\textit{Multi-Head Attention (MHA).} The structure of MHA is shown in  Figure~\ref{fig:MHA}.
The number of channels of \textbf{Q} and \textbf{K} is half of \textbf{V} to reduce the amount of calculation.
Following the LeViT, we use the attention bias as a relative position encoding rather than the absolute position encoding. We generate the attention bias in the way of our dual-image position encoding, and the details will be introduced later. The mechanism of MHA can be summarized as:

\vspace{-6mm}
\begin{equation}
\label{eq-m-h-a}
\begin{split}
     {\rm{Attn}}(\mathbf{Q},\mathbf{K},\mathbf{V},\mathbf{B}_i)
     = {\rm{softmax}}(\frac{\mathbf{Q}\mathbf{K}^\top}{\sqrt{d_k}}+\mathbf{B}_i)\mathbf{V}, \\
     {\mathbf{H}_i}={\rm{Hardswish}(\rm{Attn}}(\mathbf{X}\mathbf{W}_i^Q,\mathbf{X}\mathbf{W}_i^K,\mathbf{X}\mathbf{W}_i^V,\mathbf{B}_i)),\\
     {\rm{MultiHead}}(\mathbf{X}) = {\rm{Concat}}({\mathbf{H}_1},...,{\mathbf{H}_{N}}){\mathbf{W}^O}, 
\end{split}
\end{equation}
where $\mathbf{X}\in\mathbb{R}^{HW \times C}$ is the input, $\mathbf{B}_i\in\mathbb{R}^{HW \times HW}$ is the attention bias, and $\mathbf{W}_i^Q \in \mathbb{R}^{C \times D}$, $\mathbf{W}_i^K \in \mathbb{R}^{C \times D}$, $\mathbf{W}_i^V \in \mathbb{R}^{C \times 2D}$, and $\mathbf{W}^O \in \mathbb{R}^{2ND \times C}$ are parameter matrices.

\ignore{
Following LeViT, we fuse the BNs into their preceding linear layer during inference, which can speed up the model. Table~\ref{tab-fuse} shows the influence of this mechanism.
}

\ignore{
\begin{table}[t]\Huge
\begin{center}
\resizebox{0.45\textwidth}{!}{
\setlength{\tabcolsep}{3.5mm}{
\begin{tabular}{c | cc c}
\toprule
 FB
&GPU Speed ($fps$)
&CPU Speed ($fps$)
&AGX Speed ($fps$)\\
\midrule[0.5pt]
\checkmark &175 &33 &61\\
$\times$ &90 &22 &45\\
\bottomrule
\end{tabular}}}
\end{center}
\vspace{-4mm}
\caption{Comparison of speeds with and without FB. FB means fusing batch normalization into their preceding linear layer.}
\vspace{-3mm}
\label{tab-fuse}
\end{table}}

\textit{Shrink Attention (SA).} The structure of SA is shown in Figure ~\ref{fig:Shrink}. The SA connects the stages of the hierarchical transformer and downsamples the features. 
The architecture of SA is the same as MHA except for the following modifications:
1) To generate the Query \textbf{Q}, we split the 2D input features into template features (denoted as T) and search region features (denoted as S) based on their location. We reshape them to 3D features and subsample them by a factor of 2 in each spatial direction. Then we re-flatten the features and concatenate them in the spatial dimension. In this way, the size of Q is down-sampled by a factor of 4 in total, thus the final output of SA is also down-sampled. 
2) The number of channels of \textbf{V} is doubled to alleviate the information loss caused by downsampling, and the number of channels of output features is also increased.

\begin{figure}[t]
	\centering
 	\subfloat[Previous Position Encoding]{\label{fig:PE-previous}\includegraphics[width=0.9\linewidth]{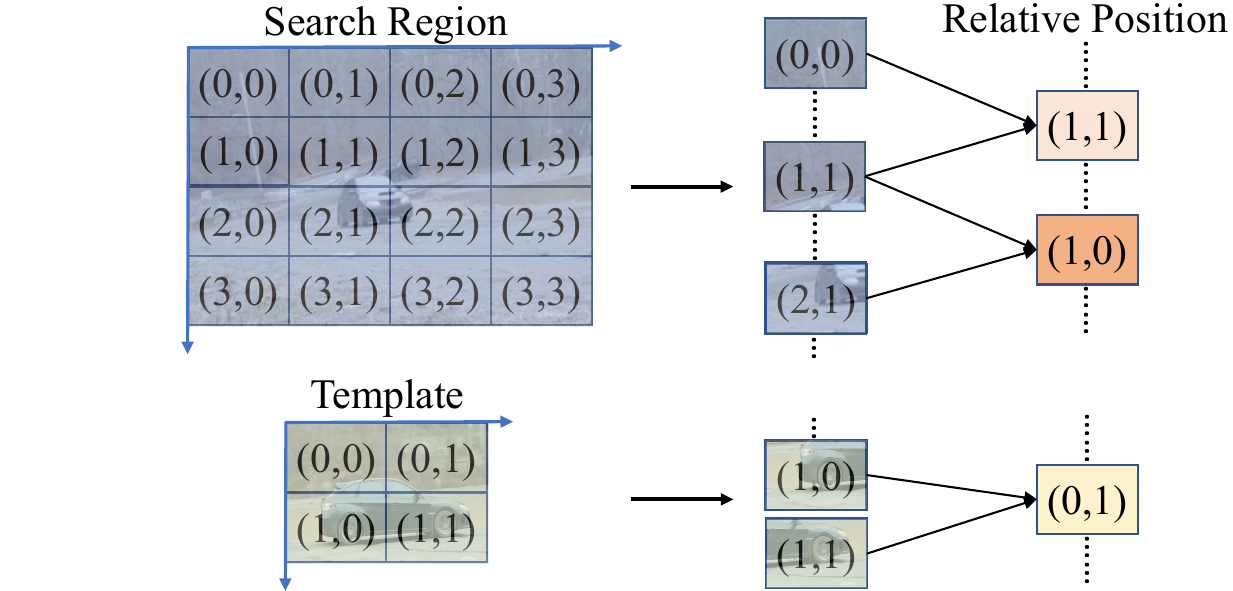}}\\	
	\vspace{-1mm}
	\subfloat[Our dual-image Position Encoding]{\label{fig:PE-our}\includegraphics[width=0.9\linewidth]{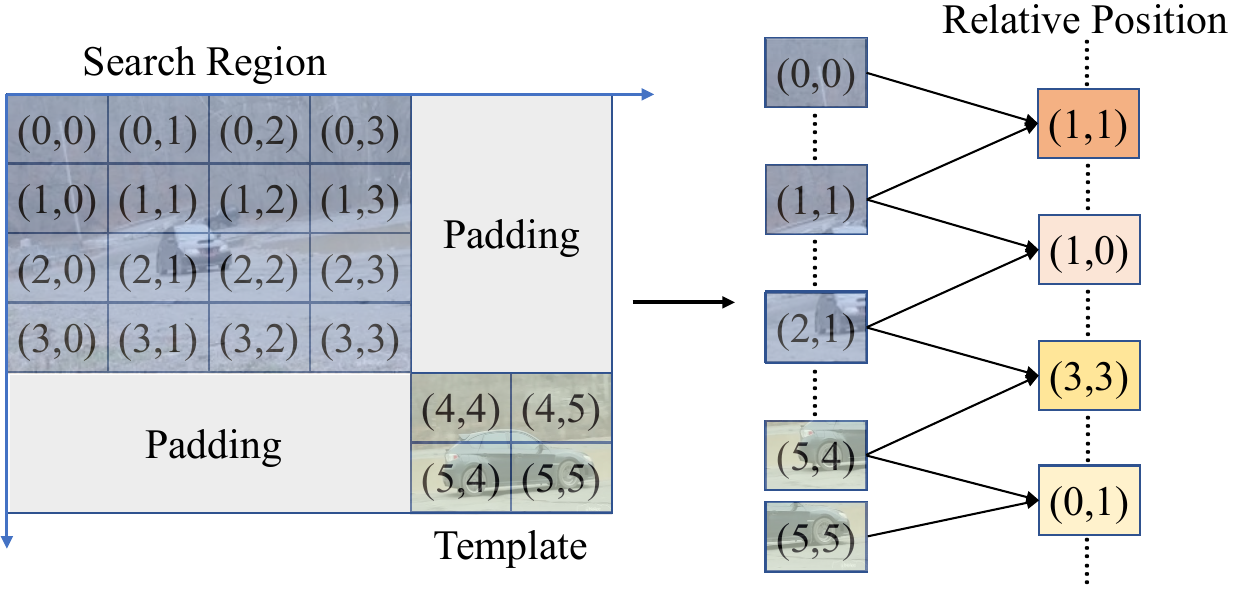}}\\
	\vspace{-2mm}
	\caption{Comparison of our dual-image position encoding and the previous position encoding.}
	\vspace{-6mm}
\end{figure}

\label{sec:PE}
\textit{Dual-image Position Encoding.} Following LeViT, we use attention bias to inject the relative position information into attention maps. To better encode the joint position information of the template and the search region, we generate the attention bias in the way of our dual-image position encoding.
Specifically, attention bias is a set of parameters that can be learned. We calculate the relative positions between every two pixels, use the relative positions as indexes to find the corresponding learned parameters, and add them to the attention map to introduce the position information. It is calculated as
\begin{equation}
\label{pos_encode}
\begin{array}{cc}
     {\rm{Bias}}^h = {\mathbf{B}^{h}(\lvert {x}-{x}^{'} \rvert, \lvert {y}-{y}^{'} \rvert)} 
\end{array}, 
\end{equation}
where $(x,y)$ and $({x}^{'},{y}^{'})$ $\in [H] \times [W]$ are the two pixels on the feature map. $\mathbf{B}^{h}$ is the learned parameters, and ${\rm{Bias}}^{h}$ is the indexed learned parameters. As shown in Figure~\ref{fig:PE-previous}, the previous position encoding encodes the template and the search region separately, and the positions of the two images partially overlap, causing information confusion. More concretely, the position of the template and the upper left portion of the search region is the same.
To address this problem, in our dual-image position encoding, we diagonally arrange the template and the search region and encode their position information jointly as shown in Figure~\ref{fig:PE-our}. The diagonal arrangement encodes unique horizontal and vertical coordinates for each pixel of the template and search region, avoiding the confusion of detailed position information.

\subsection{Bridge Module and Head}
\textit{Bridge Module.} The Bridge Module fuses features of different stages of the hierarchical transformer to obtain the enhanced feature that contains rich detailed and semantic information. It bridges the lightweight hierarchical transformer and the tracking framework. 
To ensure the efficiency of the model, we expect the Bridge Module to be a minimal architecture, that is, it should be as concise as possible while being effective. To this end, we employ a very simple architecture for the Bridge Module and find it provides compelling results.
As shown in the red box in Figure~\ref{fig:framework}, the transformer outputs three 2D features with different sizes.
We reshape these 2D features to 3D feature maps, denoted as $\mathbf{S_{min}}$, $\mathbf{S_{mid}}$, and $\mathbf{S_{max}}$. First, we upsample $\mathbf{S_{min}}$ and add it to $\mathbf{S_{mid}}$. Then, we upsample the obtained feature and add it to $\mathbf{S_{max}}$, getting the final enhanced feature. We employ a transpose convolutional layer with stride 2 for all upsampling.
The mechanism of the Bridge Module can be summarized as 
\begin{equation}
\begin{split}
    {\mathbf{O_{s}}}
    = {\mathbf{S_{max}}}+{\rm{Upsample}}({\mathbf{S_{mid}}}+{\rm{Upsample}}({\mathbf{S_{min}}})) 
\end{split}
\label{eq-att}
\end{equation}
where $\mathbf{O_{s}} \in {\mathbb{R}}^{{H_{max}} \times W_{max} \times C_{max}}$ is the output of the Bridge Module; $\mathbf{S_{max}} \in {\mathbb{R}}^{{H_{max}} \times W_{max} \times C_{max}}$, $\mathbf{S_{mid}} \in {\mathbb{R}}^{{H_{mid}} \times W_{mid} \times C_{mid}}$ and $\mathbf{S_{min}} \in {\mathbb{R}}^{{H_{min}} \times W_{min} \times C_{min}}$ are feature maps output by the lightweight hierarchical transformer. The Bridge Module combines the deep semantic information and the shallow detail information, alleviating the information loss caused by the large-stride downsampling. This
minimal network provides compelling results while remaining efficient.

\textit{Head.} We employ the corner head~\cite{Stark} for prediction. First, we calculate the attention map between ${\mathbf{G}}$ and ${\mathbf{O_{s}}}$. Then, we re-weight ${\mathbf{O_{s}}}$ with the attention map. In this way, the local features are enhanced or suppressed according to global information. Finally, ${\mathbf{O_{s}}}$ is fed to a fully-convolution network, obtaining the coordinates of the target.

\begin{table}[t]\footnotesize
  \centering
  \setlength{\tabcolsep}{2mm}{  
  \begin{tabular}{c | l | c c c c c c}
    \toprule
     \multicolumn{2}{c|}{Model}&HiT-Base &&HiT-Small &&HiT-Tiny\\
    \midrule[0.5pt]
     \multirow{2}*{{PyTorch}}&GPU &175&&192&&204\\
    \multirow{2}*{{Speed (\emph{fps})}}&CPU &33&&72&&76\\
    &AGX &61&&68&&77\\
    \midrule[0.5pt]
    \multirow{2}*{{ONNX}}&GPU &274&&400&&455\\
    \multirow{2}*{{Speed (\emph{fps})}}&CPU &42&&98&&125\\
    &AGX &75&&119&&145\\
    \midrule[0.5pt]
    \multicolumn{2}{c|}{Macs(G)} &4.34&&1.13&&0.99\\
    \midrule[0.5pt]
    \multicolumn{2}{c|}{Params(M)} &42.14&&11.03&&9.59\\
  \bottomrule
\end{tabular}
}
\vspace{-2mm}
  \caption{Details of our HiT model variants.}
  \label{tab-modelf}
\vspace{-5mm}
\end{table}

\begin{table*}[t]\footnotesize
  \centering
  \caption{State-of-the-art comparison on TrackingNet~\cite{trackingnet}, LaSOT~\cite{LaSOT}, and GOT-10k~\cite{GOT10K} benchmarks. We use \textcolor{gray}{gray} color to denote our trackers. The best three real-time results are shown in \textbf{\textcolor{cRed}{red}}, \textcolor{blue}{blue} and \textcolor{cGreen}{green} fonts, and the best non-real-time results are shown in \uline{\emph{underline}} font.}
  \label{tab-sota}
  \vspace{-2mm}
  \setlength{\tabcolsep}{2mm}{  
  \resizebox{\textwidth}{!}{
  \begin{tabular}{c|l|c ccc c ccc c ccc c ccc}
    \toprule
    & \multirow{2}*{Method} &  \multicolumn{3}{c}{TrackingNet}&& \multicolumn{3}{c}{LaSOT} &&  \multicolumn{3}{c}{GOT-10k} && \multicolumn{3}{c}{PyTorch Speed (\emph{fps})} \\
    \cline{3-5}
    \cline{7-9}
    \cline{11-13}
    \cline{15-17}
   && AUC&P$_{Norm}$&P && AUC&P$_{Norm}$&P  && AO&SR$_{0.5}$&SR$_{0.75}$ && GPU&CPU&AGX\\
    \midrule[0.5pt]
    \multirow{9}*{\rotatebox{90}{Real-time}}&\cellcolor{gray!15}HiT-Base&\cellcolor{gray!15}\textbf{\textcolor{cRed}{80.0}}&\cellcolor{gray!15}\textbf{\textcolor{cRed}{84.4}}&\cellcolor{gray!15}\textbf{\textcolor{cRed}{77.3}} &\cellcolor{gray!15} &\cellcolor{gray!15}\textbf{\textcolor{cRed}{64.6}}&\cellcolor{gray!15}\textbf{\textcolor{cRed}{73.3}}&\cellcolor{gray!15}\textbf{\textcolor{cRed}{68.1}} &\cellcolor{gray!15}&\cellcolor{gray!15}\textcolor{blue}{64.0}&\cellcolor{gray!15}\textcolor{cGreen}{72.1}&\cellcolor{gray!15}\textbf{\textcolor{cRed}{58.1}} &\cellcolor{gray!15}&\cellcolor{gray!15}175&\cellcolor{gray!15}33&\cellcolor{gray!15}61\\
    &\cellcolor{gray!15}HiT-Small &\cellcolor{gray!15}\textcolor{blue}{77.7} &\cellcolor{gray!15}\textcolor{cGreen}{81.9} &\cellcolor{gray!15}\textcolor{blue}{73.1} &\cellcolor{gray!15} &\cellcolor{gray!15}\textcolor{blue}{60.5} &\cellcolor{gray!15}\textcolor{cGreen}{68.3} &\cellcolor{gray!15}\textcolor{blue}{61.5} &\cellcolor{gray!15} &\cellcolor{gray!15}\textcolor{cGreen}{62.6} &\cellcolor{gray!15}71.2 &\cellcolor{gray!15}\textcolor{cGreen}{54.4} &\cellcolor{gray!15} &\cellcolor{gray!15}192 &\cellcolor{gray!15}72 &\cellcolor{gray!15}68  \\
    &\cellcolor{gray!15}HiT-Tiny &\cellcolor{gray!15}74.6 &\cellcolor{gray!15}78.1 &\cellcolor{gray!15}68.8 &\cellcolor{gray!15} &\cellcolor{gray!15}54.8 &\cellcolor{gray!15}60.5 &\cellcolor{gray!15}52.9 &\cellcolor{gray!15} &\cellcolor{gray!15}52.6 &\cellcolor{gray!15}59.3 &\cellcolor{gray!15}42.7 &\cellcolor{gray!15} &\cellcolor{gray!15}204 &\cellcolor{gray!15}76 &\cellcolor{gray!15}77  \\
    %\cline{2-17}
    &FEAR~\cite{borsuk2022fear}\footnotemark[1]& -&-&- && 53.5&-&54.5 &&  61.9&\textcolor{blue}{72.2}&- && 105&60&38\\
    &HCAT~\cite{chen2022efficient}& \textcolor{cGreen}{76.6}&\textcolor{blue}{82.6}&\textcolor{cGreen}{72.9} && \textcolor{cGreen}{59.3}&\textcolor{blue}{68.7}&\textcolor{cGreen}{61.0} &&  \textbf{\textcolor{cRed}{65.1}}&\textbf{\textcolor{cRed}{76.5}}&\textcolor{blue}{56.7} &&195&45&55 \\
    &E.T.Track~\cite{ETTrack}& 75.0&80.3&70.6 && 59.1&-&- &&  -&-&- &&40&47&20\\
    &LightTrack~\cite{yan2021lighttrack}& 72.5&77.8&69.5 && 53.8&-&53.7 &&  61.1&71.0&- && 128&41&36\\
    &ATOM~\cite{ATOM}& 70.3&77.1&64.8 && 51.5&57.6&50.5 &&  55.6&63.4&40.2 &&83&18&22 \\
    &ECO~\cite{danelljan2017eco}& 55.4&61.8&49.2 && 32.4&33.8&30.1 &&  31.6&30.9&11.1 && 240&15&39 \\
    \midrule[0.5pt]
    \multirow{10}*{\rotatebox{90}{Non-real-time}}&OSTrack-256~\cite{ostrack}& 83.1&87.8&82.0 && 69.1&78.7&75.2 &&  71.0&80.4&68.2 && 105&11&19 \\
    &MixFormer-L~\cite{mixformer}& \emph{\uline{83.9}}&\emph{\uline{88.9}}&83.1 && \uline{\emph{70.1}}&\uline{\emph{79.9}}&\uline{\emph{76.3}} &&  \uline{\emph{75.6}}&\uline{\emph{85.7}}&\uline{\emph{72.8}} &&18&-&-\\
    &Sim-B/16~\cite{simtrack}&82.3&-&\uline{\emph{86.5}}&&69.3&78.5&-&&68.6&78.9&62.4&&87&10&16\\
    &STARK-ST50~\cite{Stark}& 81.3&86.1&-  && 66.6&-&- && 68.0&77.7&62.3 && 50&7&13 \\
    &TransT~\cite{TransT}& 81.4&86.7&80.3 && 64.9&73.8&69.0 && 72.3&82.4&68.2 && 63&5&13  \\
    &TrDiMP~\cite{wang2021transformer}& 78.4&83.3&73.1 && 63.9&-& 61.4 &&  68.8&80.5&59.7 && 41&5&10 \\
    &TrSiam~\cite{wang2021transformer}& 78.1&82.9&72.7 && 62.4&-&60.6 &&  67.3&78.7&58.6 && 40&5&10 \\
    &PrDiMP~\cite{PrDiMP}& 75.8&81.6&70.4 && 59.8&68.8&60.8 &&  63.4&73.8&54.3 &&47&6&11 \\
    &DiMP~\cite{DiMP}& 74.0&80.1&68.7 && 56.9&65.0&56.7 && 61.1&71.7&49.2 && 77 &10&17  \\
    &SiamRPN++~\cite{SiamRPNplusplus}& 73.3&80.0&69.4 && 49.6&56.9&49.1 && 51.7&61.6&32.5 && 56 &4&11 \\
\bottomrule
\end{tabular}
}}
\vspace{-2.5mm}
\end{table*}

\subsection{Training objective and Inference}
\label{sec:training_and_inference}
We combine the $\ell_1$ loss and the generalized GIoU loss~\cite{GIoU} as the training objective. The loss function can be formulated as
\begin{equation}
	\label{equ-loss-loc}
	\begin{aligned}
		\mathcal{L}=\lambda_{G}\mathcal{L}_{GIoU}(b_i,{\hat{b}}_i)+\lambda_{l}\mathcal{L}_l(b_i,\hat{b}_i).
	\end{aligned}
\end{equation}
where $b_i$ represents the groundtruth, and ${\hat{b}}_i$ represents the predicted box. $\lambda_{G}$ and $\lambda_{l}$ are weights, in experiments, we set $\lambda_{G}=2$ and $\lambda_{l}=5$. During inference, the template is initialized in the first frame of a video sequence. %We do not use the online template update mechanism~\cite{Stark}. 
For each subsequent frame, the search region is cropped based on the target's bounding box of the previous frame.
The whole framework is end-to-end. The template and search images are input into our tracker, and the output of the model is the final result. We do not use any additional post-processing methods, such as window penalty and scale penalty~\cite{SiameseRPN}. 

\section{Experiments}
\label{sec:Experiments}

\subsection{Implementation Details}

\textit{Model.} We develop three variants of HiT models with different lightweight transformers, as elaborated in Tab.~\ref{tab-modelf}. We adopt LeViT-384~\cite{graham2021levit}, LeViT-128, and LeViT-128S for HiT-Base, HiT-Small, and HiT-Tiny, respectively. 
In addition, Tab.~\ref{tab-modelf} reports model parameters, FLOPs, and inference speed on multiple devices. All the models are implemented with Python 3.8 and PyTorch 1.11.0. 

\textit{Training.} The training datasets for our model include the train-splits of TrackingNet~\cite{trackingnet}, GOT-10k~\cite{GOT10K}, LaSOT~\cite{LaSOT}, and COCO2017~\cite{COCO}. The input of the network is an image pair consisting of a template image and a search image. For video datasets, we sample the image pair from a random video sequence. For the image dataset COCO, we randomly select an image and apply data augmentations to generate an image pair. Common data augmentations such as scaling, translation, and jittering are applied on the image pair. The search region and the template are obtained by expanding the target box by a factor of $4$ and $2$, respectively. The search and template images are resized to $256 \times 256$ and $128 \times 128$, respectively. The transformer is initialized with ImageNet~\cite{ImageNet} pretrained LeViT~\cite{graham2021levit}, and the other parameters of HiT are initialized randomly. The optimizer is the AdamW optimizer~\cite{AdamW}, with the weight decay of 1e-4. The initial learning rate of HiT is 5e-4. We use 4 Nvidia RTX 3090 GPUs to train our model for 1500 epochs with a batch size of 128. Each epoch contains 60,000 sampling pairs. The learning rate is reduced by $10\times$ at epoch 1200.
%%%%%%%%%%%%

\textit{Inference.} As stated in Sec.~\ref{sec:training_and_inference}, the HiT framework is end-to-end, and we do not involve any hyper-parameters during inference.

\footnotetext[1]{Due to limitations in access to the FEAR-L model, we compare our method with FEAR-XS in this table. Nevertheless, our model also performs better than FEAR-L. For example, HiT-Base performs 6.7\% higher than FEAR-L (in AUC) on LaSOT.}
%Additionally, our model runs faster than FEAR-XS and is, therefore, faster than FEAR-L as well, since FEAR-L runs 5.4 times slower than FEAR-XS according to the FEAR paper.

\subsection{State-of-the-art Comparisons} According to the speed on edge device Nvidia Jetson AGX Xavier, we divide trackers into real-time trackers and non-real-time trackers. Following the VOT real-time setting~\cite{vot2020}, we set the real-time line at 20 \emph{fps}.
We compare HiT with the state-of-the-art real-time trackers and non-real-time trackers on six tracking benchmarks. We evaluate these trackers' speed on three platforms: Nvidia GeForce RTX 2080 GPU, Intel Core i9-9900K @ 3.60GHz CPU, and Nvidia Jetson AGX Xavier edge device. 
Tables~\ref{tab-sota} and~\ref{tab-sota-small} show the results.

\textit{TrackingNet.} TrackingNet~\cite{trackingnet} is a large-scale dataset containing a variety of situations in natural scenes and multiple categories, and its test set includes 511 video sequences. As reported in Table~\ref{tab-sota}, HiT-Base and HiT-small achieve competitive results compared with the previous real-time trackers. HiT-Base gets the best AUC of 80.0\%, surpassing the previous best real-time tracker HCAT~\cite{chen2022efficient} by 3.4\%. Compared to non-real-time tracker STARK-ST50~\cite{Stark}, HiT-Base achieves comparable performance to it in AUC (80.0 $vs$. 81.3) while being $3.5 \times$ faster on the GPU, $4.7 \times$ faster on the CPU, and $4.7 \times$ faster on the AGX.

\textit{LaSOT.} LaSOT~\cite{LaSOT} is a large-scale, long-term dataset containing 1400 video sequences, with 1120 training videos and 280 test videos. The results on LaSOT are shown in Table~\ref{tab-sota}. HiT-Base achieves the best real-time results of 64.6\%, 73.3\%, and 68.1\% in AUC, P$_{Norm}$, and P, respectively. HiT-Small achieves the second-best AUC score. Compared with the recent efficient tracker FEAR~\cite{borsuk2022fear}, HiT-Base and HiT-Small outperform it by 11.1\% and 7.0\% in AUC. Moreover, HiT-Base and HiT-Small surpass the third-best real-time tracker HCAT~\cite{chen2022efficient} by 5.3\% and 1.2\% in AUC. Compared with the non-real-time tracker TransT~\cite{TransT}, HiT-Base performs only 0.3\% lower but has a much faster speed.

\textit{GOT-10k.} GOT-10k~\cite{GOT10K} is a large-scale and challenging dataset that contains 10k training sequences and 180 test sequences. As shown in Table~\ref{tab-sota}, HiT-Base obtains the second best real-time results of 64.0\% AO score. HiT-Small obtains the third-best AO score of 62.6\%. HiT-Base surpasses the recent efficient tracker FEAR~\cite{borsuk2022fear} by 2.1\% . 

\begin{table}[t]\small
  \centering
  \setlength{\tabcolsep}{4mm}{
  \resizebox{1\linewidth}{!}{
    \begin{tabular}{c|lccc}
    \toprule
    &Method&NFS&UAV123&LaSOT$_{ext}$\\
    \midrule[0.5pt]
    \multirow{9}*{\rotatebox{90}{Real-time}} &\cellcolor{gray!15}HiT-Base &\cellcolor{gray!15}\textbf{\textcolor{cRed}{63.6}} &\cellcolor{gray!15}\textbf{\textcolor{cRed}{65.6}} &\cellcolor{gray!15}\textbf{\textcolor{cRed}{44.1}}\\
    &\cellcolor{gray!15}HiT-Small &\cellcolor{gray!15}\textcolor{cGreen}{61.8} &\cellcolor{gray!15}\textcolor{cGreen}{63.3} &\cellcolor{gray!15}\textcolor{blue}{40.4}\\
    &\cellcolor{gray!15}HiT-Tiny &\cellcolor{gray!15}53.2 &\cellcolor{gray!15}58.7 &\cellcolor{gray!15}35.8\\
    &HCAT~\cite{chen2022efficient}&\textcolor{blue}{63.5}&62.7&-\\
    &FEAR~\cite{borsuk2022fear}&61.4&-&-\\
    &E.T.Track~\cite{ETTrack}&59.0&62.3&-\\
    &LightTrack~\cite{yan2021lighttrack}&55.3&62.5&-\\
    &ATOM~\cite{ATOM}&58.4&\textcolor{blue}{64.2}&\textcolor{cGreen}{37.6}\\ 
    &ECO~\cite{danelljan2017eco}&46.6&53.2&22.0\\
    \midrule[0.5pt]
    \multirow{7}*{\rotatebox{90}{Non-real-time}}&OSTrack-256~\cite{ostrack}&64.7&68.3&\emph{\uline{47.4}}\\
    &TransT~\cite{TransT}&65.7&\emph{\uline{69.1}}&-\\
    &TrDiMP~\cite{wang2021transformer}&\emph{\uline{66.5}}&67.5&-\\ 
    &TrSiam~\cite{wang2021transformer}&65.8&67.4&-\\
    &PrDiMP~\cite{PrDiMP}&63.5&68.0&-\\ 
    &DiMP~\cite{DiMP} &62.0&65.3&39.2\\
    &SiamRPN++~\cite{SiamRPNplusplus}&50.2&61.6&34.0\\
    \bottomrule
    \end{tabular}
  }}
  \vspace{-2mm}
    \caption{Comparison with state-of-the-art methods on additional benchmarks in AUC score.}
  \label{tab-sota-small}
\vspace{-2mm}
\end{table}

\begin{table}[h]\tiny
\centering
\resizebox{1.0\linewidth}{!}{
    \begin{tabular}{l|ccc}
    \toprule[0.5pt]
    Method &EAO&Accuracy&Robustness\\
    \midrule[0.5pt]
    HiT-Base &\textcolor{cRed}{0.252}&\textcolor{cRed}{0.447}&\textcolor{cRed}{0.688} \\
    \midrule[0.1pt]
    FEAR ~\cite{borsuk2022fear} &\textcolor{blue}{0.250} &\textcolor{blue}{0.436} &\textcolor{blue}{0.655} \\
    STARK-S~\cite{Stark} &\textcolor{cGreen}{0.237} &\textcolor{cGreen}{0.407} &0.631 \\
    STARK-Lightning~\cite{Stark} &{0.204} &{0.391} &0.565 \\
    LightTrack~\cite{yan2021lighttrack} &0.225 &0.391 &\textcolor{cGreen}{0.641} \\
    E.T.Track~\cite{ETTrack} &0.224&0.372&0.631\\
    \bottomrule[0.5pt]
    \end{tabular}
  }
  \vspace{-2mm}
\caption{VOT real-time experiment on NVidia Jetson AGX.}
\label{tab-vot}
\vspace{-3mm}
\end{table}

\textit{Speed.} Table~\ref{tab-sota} reports the speeds of trackers. On the GPU, HiT-Base, HiT-Small, and HiT-Tiny run at 175 \emph{fps}, 192 \emph{fps}, and 204 \emph{fps}, which are 1.66$\times$, 1.82$\times$, and 1.94$\times$ faster than FEAR~\cite{borsuk2022fear}. 
On the AGX edge device, HiT-Base, HiT-Small, and HiT-Tiny run at 61 \emph{fps}, 68 \emph{fps}, and 77 \emph{fps}, which are 1.61$\times$, 1.79$\times$, and 2.03$\times$ faster than FEAR.
On the CPU, HiT-Base, HiT-Small, and HiT-Tiny run at 33 \emph{fps}, 72 \emph{fps}, and 76 \emph{fps}. Only HiT-base is slower than FEAR but still achieves real-time speed.
Overall, HiT achieves fast speeds on multiple devices. We believe that the fast speed is beneficial for the applicability of tracking.

\textit{NFS.} NFS~\cite{NFS} is a challenging dataset with fast-moving objects, which includes 100 video sequences. Table~\ref{tab-sota-small} shows that HiT-Base and HiT-Small achieve the best and the third-best real-time performance, respectively.

\textit{UAV123.} The UAV123 dataset~\cite{UAV} is constructed with low-altitude UAVs and contains 123 video clips. As shown in Table~\ref{tab-sota-small}, HiT-Base achieves the best results compared to the other real-time trackers, achieving an AUC score of 65.6\%. HiT-Base performs superior to HCAT~\cite{chen2022efficient} and E.T.Track~\cite{ETTrack} with 2.9\% and 3.3\%, respectively.

\textit{LaSOT$_{ext}$.} LaSOT$_{ext}$~\cite{lasot_journal} is a recently released tracking dataset consisting of 150 videos from 15 object classes. It is an extension of LaSOT. The results of HiT on LaSOT$_{ext}$ are shown in Table~\ref{tab-sota-small}. HiT-Base, HiT-Small, and HiT-Tiny achieve competitive results with 44.1\%, 40.4\%, and 35.8\% AUC scores, respectively.

\textit{VOT.} We also conduct VOT real-time experiments on NVidia Jetson AGX  using the VOT2021 benchmark~\cite{vot2021}. The results are shown in the Table~\ref{tab-vot}. HiT-Base achieves the best results compared to the other real-time trackers, achieving an EAO score of 25.2\%.
\subsection{Ablation Study and Analysis}
In this section, we provide detailed ablation experiments to analyze our HiT method. For the ablation study, we use HiT-Base as the baseline model. All models in ablation experiments are trained for 500 epochs.

\begin{table}[t]\scriptsize
  \centering
  \setlength{\tabcolsep}{2mm}{
  \resizebox{\linewidth}{!}{
    \begin{tabular}{c|ccc|ccc}
    \toprule
    \#&Max&Mid&Min&LaSOT&TrackingNet&GOT-10k\\
    \midrule[0.5pt] 
    \rowcolor{gray!15}
    1&\checkmark&\checkmark&\checkmark&\textbf{\textcolor{cRed}{63.7}}&\textbf{\textcolor{cRed}{78.9}}&\textbf{\textcolor{cRed}{65.4}}\\
    2&\checkmark&&&62.1&78.3&63.4\\
    3&&\checkmark&&61.9&78.1&64.1\\
    4&&&\checkmark&57.9&73.0&62.1\\
    5&\checkmark&\checkmark&&62.6&78.7&63.0\\
    6&\checkmark&&\checkmark&58.8&77.2&60.4\\
    7&&\checkmark&\checkmark&60.3&78.4&63.6\\
    \bottomrule
    \end{tabular}
  }}
 \vspace{-2mm}
   \caption{Comparison with different feature combining manners in AUC. We use \textcolor{gray}{gray} color to denote the default setting. The best results are shown in the \textbf{\textcolor{cRed}{red}} fonts. Max, Mid, and Min denote the features of the transformer's first, second, and third stages.}
  \label{tab-components}
   \vspace{-3mm}
\end{table}

\textit{Different combinations of features.} 
To verify the effectiveness of the Bridge Module and explore which features are important, we compare different feature combinations in the Bridge Module.
Table~\ref{tab-components} shows the results. Max, Mid, and Min denote the features of the transformer's first, second, and third stages, respectively. For a fair comparison, the features are upsampled to the same resolution in the comparison. The first row (\#1) is our default setting. First, we do not use our Bridge Module and make predictions on the independent Max, Mid and Min features.
Table~\ref{tab-components} (\#2, \#3, and \#4) shows these methods lead to inferior results, demonstrating the effectiveness of feature fusing with our Bridge Module. Second, Table~\ref{tab-components} (\#5, \#6, and \#7) reports the results of other candidate combination manners, and our default method works best. In our default method, using all three features bring more semantic and detailed information, leading to better results. 

\begin{figure}[!t]
\begin{center}
\includegraphics[width=0.95\linewidth]{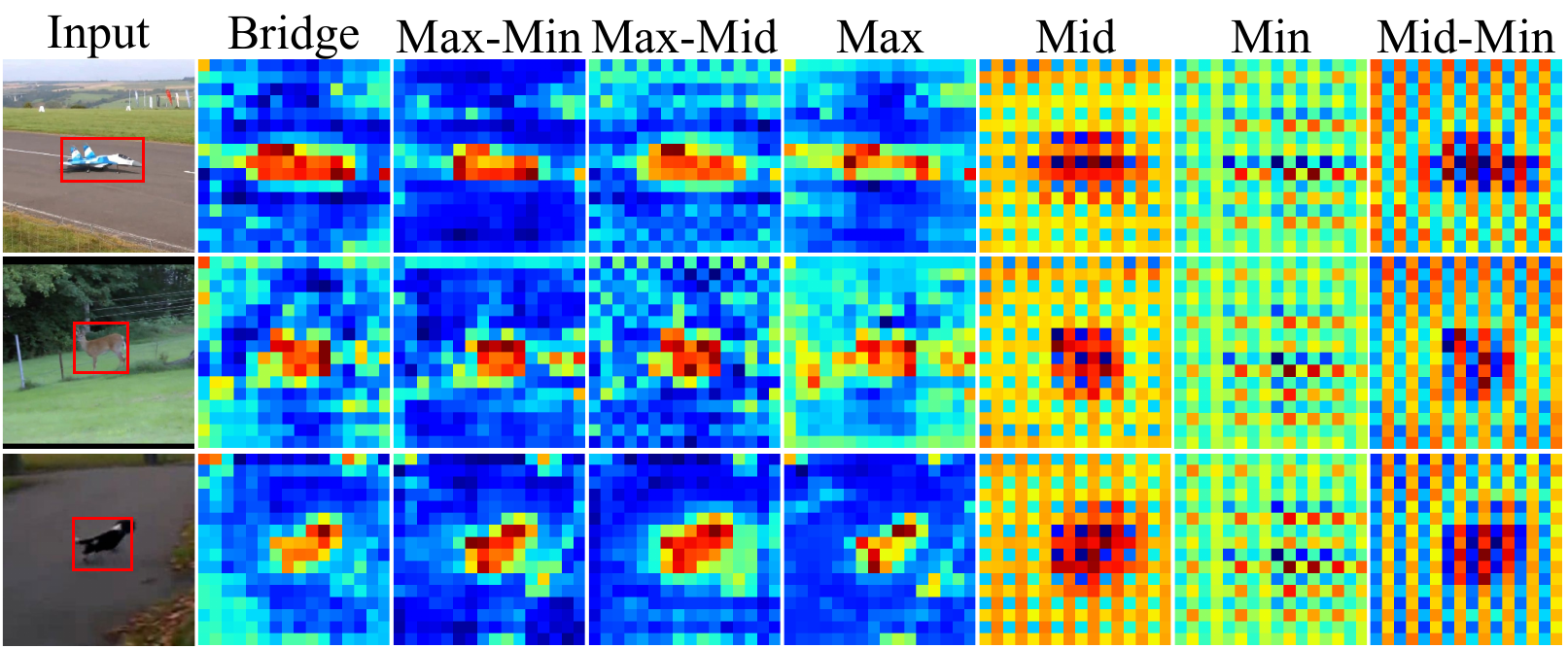}
\end{center}
    \vspace{-2mm}
   \caption{Visualization of the attention maps in the corner head of different feature combining manners. Bridge means our default manner, Max-Min means combining the Max and the Min features, Max-Mid means combining the Max and the Min features, Max, Mid, and Min mean only using the Max feature, Mid feature, and Min feature, respectively.}
\label{fig:attn_vis}
\vspace{-3mm}
\end{figure}

To further understand the Bridge Module, we visualize the attention map in the corner head of these features combining manners in Figure~\ref{fig:attn_vis}. In the visualization results, first, we find a collapse phenomenon in the methods that do not use the Max feature. Taking the Mid manner as an example, the final feature is from the second stage of the transformer, and it is up-sampled by a factor of 2. In this way, one pixel on the feature map is up-sampled to four pixel points. In the visualization result, we can see that the attention collapses to a relatively fixed distribution for every four upsampling grids. The Min column and the Mid-Min column are similar to the Mid column. This shows that even if the deep feature is up-sampled to a larger resolution, it does not bring more detailed information. Therefore, it is crucial to involve the shallow large-resolution feature to supplement the information. Second, we find the attention map of our default method is more accurate than the methods that do not use the Min feature. This demonstrates that using deep features to supplement semantic information helps to improve the discriminative ability.

\vspace{0mm}
\begin{table}[t]
\begin{center}
\resizebox{\linewidth}{!}{
\setlength{\tabcolsep}{6mm}{
\begin{tabular}{l|l|cccc}
\toprule
\#&PE &LaSOT	&TrackingNet	&GOT-10k\\
\midrule[0.5pt]
\rowcolor{gray!15}
1&DI  &\textbf{\textcolor{cRed}{63.7}}	 &\textbf{\textcolor{cRed}{78.9}} &\textbf{\textcolor{cRed}{65.4}}\\
2&Abs &60.2 &77.2 &61.2\\
3&Sep &62.4 &77.6 &63.1\\
4&Ver &61.1 &78.4 &63.5\\
5&Hor &61.0 &78.5 &63.7\\
\bottomrule
\end{tabular}}}
\vspace{-2mm}
\caption{Comparison of different Position Encoding (PE) in AUC score. DI denotes our dual-image PE. Abs denotes the absolute PE. Sep denotes the relative PE which encodes the template and search region separately. Ver and Hor denote the joint encoding of the template and search images in a vertical and horizontal arrangement, respectively.
}
\label{ablation on position encoding}
\vspace{-10mm}
\end{center}
\end{table}

\textit{Different Position Encoding.} Previous transformer-based trackers~\cite{TransT,Stark} encode the position information of the search image and the template image separately. In our dual-image position encoding method, we assign a unique position for each image and jointly encode their position information. Here, we compare our method with four potential encoding methods, and the results are reported in Table~\ref{ablation on position encoding}. First, we compare our method with the absolute position encoding (denoted as Abs) and the relative position encoding which encodes the search and template images separately (denoted as Sep). Table~\ref{ablation on position encoding} (\#1 and \#2) shows these methods perform inferior to our dual-image position encoding. The separate encoding does not model the positional relationship between the search and template images, and introduces overlapping positions of them, leading to inferior performance.
Second, in our dual-image position encoding, we also explore different arrangements of the template and search region. By default, we diagonally arrange the template and the search region, as shown in Figure~\ref{fig:PE-our}. Here we compare it with two other arrangements: the vertical arrangement (denoted as Ver) and the horizontal arrangement (denoted as Hor). Table~\ref{ablation on position encoding} (\#1, \#4 and \#5) shows the default diagonally arrangement achieves the best performance. In the vertical and horizontal arrangements, the horizontal and vertical positions of the template and the search region are overlapping, leading to information loss. The diagonal arrangement assigns unique horizontal and vertical positions for the template and the search region, which is more informative. Therefore, we choose the diagonal arrangement.

\vspace{0mm}
\begin{table}[t]
\begin{center}
\resizebox{0.95\linewidth}{!}{
\begin{tabular}{c|c|cc}
\toprule
& &LeViT-384~\cite{graham2021levit} &PVT-Small~\cite{wang2021pyramid}\\
\midrule[0.5pt]
\multirow{3}{*}{Benchmarks}&LaSOT&63.7&63.9\\
&TrackingNet&78.9&78.4\\
&GOT-10k&65.4&64.8\\
\midrule[0.5pt]
\multirow{3}{*}{PyTorch Speed (\emph{fps})}&GPU&175&91\\
&CPU&33&22\\
&AGX&61&30\\
\midrule[0.5pt]
\multirow{3}{*}{ONNX Speed (\emph{fps})}&GPU&274&133\\
&CPU&42&25\\
&AGX&75&32\\
\bottomrule
\end{tabular}}
\vspace{-3mm}
\caption{HiT with different lightweight hierarchical vision transformers.}
\label{different backbone}
\vspace{-9mm}
\end{center}
\end{table}

\textit{Different Backbones.} To evaluate the generalization of our HiT framework. We expand our framework with another hierarchical vision transformer PVT~\cite{wang2021pyramid}, the results are shown in Table~\ref{different backbone}. We employ PVT-Small\cite{wang2021pyramid} as the transformer backbone, and the other parts are consistent with HiT-Base.
From Table~\ref{different backbone}, we can see that HiT with PVT-Small obtains a 63.9\% AUC score on LaSOT, 78.4\% AUC score on TrackingNet, and 64.8\% AO score on GOT-10k, while the speed on all three platforms is real-time. This is also a competitive result compared with our base model with LeViT-384 and other efficient trackers. This demonstrates a superior generalization ability of our framework.

\vspace{-1mm}
\section{Conclusion}
\label{sec:conclusion}
\vspace{0mm}

This work proposes a new family of efficient transformer-based tracking models, named HiT. HiT alleviates the gap between the tracking framework and the lightweight hierarchical transformers through our Bridge Module and dual-image position encoding. Extensive experiments demonstrate HiT achieves promising performance compared to state-of-the-art efficient trackers while running at a very fast speed. We hope this work could facilitate the practical applicability of visual tracking and narrow the gap between the tracking and lightweight transformer research.

\textit{Limitation.} One limitation of HiT is that, despite achieving good performance, it shows the difficulty in dealing with distractors, since the method does not employ an explicit distractor-handling module. Moreover, this work focuses on bridging the gap between lightweight hierarchical transformers and the tracking framework. Therefore, we only make minimal adjustments to the existing hierarchical transformer but do not design a new transformer. In future work, we will investigate the lightweight transformer customized for tracking, and we hope this work could provide a basis for this.\\

\vspace{-3mm}
\emph{Acknowledgement.}
Dr. Wang and Dr. Lu was supported in part by National Natural Science Foundation of China (Nos.62293540, 62293542, 62022021), in part by Joint Fund of Ministry of Education for Equipment Pre-research (No.8091B032155), in part by Fundamental Research Funds for the Central Universities (No.DUT22QN228).

\clearpage

{\small
\bibliographystyle{ieee_fullname}
\bibliography{main}

\begin{thebibliography}{10}\itemsep=-1pt

\bibitem{SiameseFC}
Luca Bertinetto, Jack Valmadre, Jo{\~a}o~F Henriques, Andrea Vedaldi, and
  Philip H~S Torr.
\newblock {Fully-Convolutional Siamese Networks for Object Tracking}.
\newblock In {\em ECCV}, pages 850--865, 2016.

\bibitem{DiMP}
Goutam Bhat, Martin Danelljan, Luc~Van Gool, and Radu Timofte.
\newblock Learning {Discriminative Model Prediction for Tracking}.
\newblock In {\em ICCV}, pages 6181--6190, 2019.

\bibitem{ETTrack}
Philippe Blatter, Menelaos Kanakis, Martin Danelljan, and Luc Van~Gool.
\newblock Efficient {Visual Tracking with Exemplar Transformers}.
\newblock In {\em WACV}, pages 1571--1581, 2023.

\bibitem{borsuk2022fear}
Vasyl Borsuk, Roman Vei, Orest Kupyn, Tetiana Martyniuk, Igor Krashenyi, and
  Ji{\v{r}}i Matas.
\newblock {FEAR}: {Fast, Efficient, Accurate and Robust Visual Tracker}.
\newblock In {\em ECCV}, pages 644--663, 2022.

\bibitem{simtrack}
Boyu Chen, Peixia Li, Lei Bai, Lei Qiao, Qiuhong Shen, Bo Li, Weihao Gan, Wei
  Wu, and Wanli Ouyang.
\newblock {Backbone is All Your Need: A Simplified Architecture for Visual
  Object Tracking}.
\newblock In {\em ECCV}, pages 375--392, 2022.

\bibitem{chen2022efficient}
Xin Chen, Ben Kang, Dong Wang, Dongdong Li, and Huchuan Lu.
\newblock Efficient {Visual Tracking via Hierarchical Cross-Attention
  Transformer}.
\newblock In {\em ECCVW}, pages 461--477, 2022.

\bibitem{SeqTrack}
Xin Chen, Houwen Peng, Dong Wang, Huchuan Lu, and Han Hu.
\newblock Seqtrack: Sequence to sequence learning for visual object tracking.
\newblock In {\em CVPR}, pages 14572--14581, 2023.

\bibitem{TransT}
Xin Chen, Bin Yan, Jiawen Zhu, Dong Wang, Xiaoyun Yang, and Huchuan Lu.
\newblock Transformer {T}racking.
\newblock In {\em CVPR}, pages 8126--8135, 2021.

\bibitem{SiamBAN}
Zedu Chen, Bineng Zhong, Guorong Li, Shengping Zhang, and Rongrong Ji.
\newblock {Siamese Box Adaptive Network for Visual Tracking}.
\newblock In {\em CVPR}, pages 6667--6676, 2020.

\bibitem{mixformer}
Yutao Cui, Cheng Jiang, Limin Wang, and Gangshan Wu.
\newblock Mixformer: {End-to-End Tracking with Iterative Mixed Attention}.
\newblock In {\em CVPR}, pages 13598--13608, 2022.

\bibitem{danelljan2017eco}
Martin Danelljan, Goutam Bhat, Fahad~Shahbaz Khan, and Michael Felsberg.
\newblock {ECO}:{ Efficient Convolution Operators for Tracking}.
\newblock In {\em CVPR}, pages 6931--6939, 2017.

\bibitem{ATOM}
Martin Danelljan, Goutam Bhat, Fahad~Shahbaz Khan, and Michael Felsberg.
\newblock {ATOM: Accurate} {Tracking by Overlap Maximization}.
\newblock In {\em CVPR}, pages 4660--4669, 2019.

\bibitem{PrDiMP}
Martin Danelljan, Luc~Van Gool, and Radu Timofte.
\newblock {Probabilistic Regression for Visual Tracking}.
\newblock In {\em CVPR}, pages 7181--7190, 2020.

\bibitem{ViT}
Alexey Dosovitskiy, Lucas Beyer, Alexander Kolesnikov, Dirk Weissenborn,
  Xiaohua Zhai, Thomas Unterthiner, Mostafa Dehghani, Matthias Minderer, Georg
  Heigold, Sylvain Gelly, et~al.
\newblock {An Image is Worth 16x16 Words: Transformers for Image Recognition at
  Scale}.
\newblock In {\em ICLR}, 2021.

\bibitem{lasot_journal}
Heng Fan, Hexin Bai, Liting Lin, Fan Yang, Peng Chu, Ge Deng, Sijia Yu,
  Mingzhen Huang, Juehuan Liu, Yong Xu, et~al.
\newblock {LaSOT}: {A} {High-quality Large-scale Single Object Tracking
  Benchmark}.
\newblock {\em IJCV}, 129(2):439--461, 2021.

\bibitem{LaSOT}
Heng Fan, Liting Lin, Fan Yang, Peng Chu, Ge Deng, Sijia Yu, Hexin Bai, Yong
  Xu, Chunyuan Liao, and Haibin Ling.
\newblock {LaSOT}: {A High-Quality Benchmark for Large-Scale Single Object
  Tracking}.
\newblock In {\em CVPR}, pages 5374--5383, 2019.

\bibitem{AiATrack}
Shenyuan Gao, Chunluan Zhou, Chao Ma, Xinggang Wang, and Junsong Yuan.
\newblock {AiATrack}: Attention in attention for transformer visual tracking.
\newblock In {\em ECCV}, pages 146--164, 2022.

\bibitem{graham2021levit}
Benjamin Graham, Alaaeldin El-Nouby, Hugo Touvron, Pierre Stock, Armand Joulin,
  Herv{\'e} J{\'e}gou, and Matthijs Douze.
\newblock {LeViT: a Vision Transformer in ConvNet's Clothing for Faster
  Inference}.
\newblock In {\em ICCV}, pages 12239--12249, 2021.

\bibitem{SiamCAR}
Dongyan Guo, Jun Wang, Ying Cui, Zhenhua Wang, and Shengyong Chen.
\newblock {SiamCAR:} {S}iamese {Fully Convolutional Classification and
  Regression for Visual Tracking}.
\newblock In {\em CVPR}, pages 6268--6276, 2020.

\bibitem{ResNet}
Kaiming He, Xiangyu Zhang, Shaoqing Ren, and Jian Sun.
\newblock {Deep Residual Learning for Image Recognition}.
\newblock In {\em CVPR}, pages 770--778, 2016.

\bibitem{GOT10K}
Lianghua Huang, Xin Zhao, and Kaiqi Huang.
\newblock Got-10k: {A} {Large High-Diversity Benchmark for Generic Object
  Tracking in the Wild}.
\newblock {\em IEEE TPAMI}, 43(5):1562--1577, 2021.

\bibitem{NFS}
Hamed Kiani~Galoogahi, Ashton Fagg, Chen Huang, Deva Ramanan, and Simon Lucey.
\newblock {Need for Speed}: {A} {Benchmark for Higher Frame Rate Object
  Tracking}.
\newblock In {\em ICCV}, pages 1134--1143, 2017.

\bibitem{vot2020}
Matej Kristan, Ale{\v{s}} Leonardis, Ji{\v{r}}{\'\i} Matas, Michael Felsberg,
  Roman Pflugfelder, Joni-Kristian K{\"a}m{\"a}r{\"a}inen, Martin Danelljan,
  Luka~{\v{C}}ehovin Zajc, Alan Luke{\v{z}}i{\v{c}}, Ondrej Drbohlav, et~al.
\newblock The eighth visual object tracking {VOT}2020 challenge results.
\newblock In {\em ECCV}, pages 547--601, 2020.

\bibitem{vot2021}
Matej Kristan, Ji{\v{r}}{\'\i} Matas, Ale{\v{s}} Leonardis, Michael Felsberg,
  Roman Pflugfelder, Joni-Kristian K{\"a}m{\"a}r{\"a}inen, Hyung~Jin Chang,
  Martin Danelljan, Luka Cehovin, Alan Luke{\v{z}}i{\v{c}}, et~al.
\newblock The ninth visual object tracking vot2021 challenge results.
\newblock In {\em ICCV}, pages 2711--2738, 2021.

\bibitem{AlexNet}
Alex Krizhevsky, Ilya Sutskever, and Geoffrey~E Hinton.
\newblock Imagenet classification with deep convolutional neural networks.
\newblock In {\em NIPS}, pages 1106--1114, 2012.

\bibitem{SiamRPNplusplus}
Bo Li, Wei Wu, Qiang Wang, Fangyi Zhang, Junliang Xing, and Junjie Yan.
\newblock {SiamRPN++}: {Evolution} of {Siamese Visual Tracking with Very Deep
  Networks}.
\newblock In {\em CVPR}, pages 4282--4291, 2019.

\bibitem{SiameseRPN}
Bo Li, Junjie Yan, Wei Wu, Zheng Zhu, and Xiaolin Hu.
\newblock {High Performance Visual Tracking With Siamese Region Proposal
  Network}.
\newblock In {\em CVPR}, pages 8971--8980, 2018.

\bibitem{lin2017feature}
Tsung-Yi Lin, Piotr Doll{\'a}r, Ross Girshick, Kaiming He, Bharath Hariharan,
  and Serge Belongie.
\newblock Feature {Pyramid Networks for Object Detection}.
\newblock In {\em CVPR}, pages 936--944, 2017.

\bibitem{COCO}
Tsung-Yi Lin, Michael Maire, Serge~J. Belongie, Lubomir~D. Bourdev, Ross~B.
  Girshick, James Hays, Pietro Perona, Deva Ramanan, Piotr Doll{\'a}r, and
  C.~Lawrence Zitnick.
\newblock {Microsoft COCO}: {Common Objects in Context}.
\newblock In {\em ECCV}, pages 740--755, 2014.

\bibitem{liu2021swin}
Ze Liu, Yutong Lin, Yue Cao, Han Hu, Yixuan Wei, Zheng Zhang, Stephen Lin, and
  Baining Guo.
\newblock {Swin Transformer: Hierarchical Vision Transformer using Shifted
  Windows}.
\newblock In {\em ICCV}, pages 9992--10002, 2021.

\bibitem{AdamW}
Ilya Loshchilov and Frank Hutter.
\newblock Decoupled {Weight Decay Regularization}.
\newblock In {\em ICLR}, 2019.

\bibitem{ToMP}
Christoph Mayer, Martin Danelljan, Goutam Bhat, Matthieu Paul, Danda~Pani
  Paudel, Fisher Yu, and Luc Van~Gool.
\newblock Transforming model prediction for tracking.
\newblock In {\em CVPR}, pages 8731--8740, 2022.

\bibitem{mehta2021mobilevit}
Sachin Mehta and Mohammad Rastegari.
\newblock {MobileViT}: {Light-weight}, {General}-purpose, and {Mobile-friendly}
  {Vision Transformer}.
\newblock In {\em ICLR}, 2022.

\bibitem{UAV}
Matthias Mueller, Neil Smith, and Bernard Ghanem.
\newblock {A {Benchmark and Simulator}} for {UAV} {T}racking.
\newblock In {\em ECCV}, pages 445--461, 2016.

\bibitem{trackingnet}
Matthias Muller, Adel Bibi, Silvio Giancola, Salman Alsubaihi, and Bernard
  Ghanem.
\newblock Tracking{N}et: {A} {Large-Scale Dataset and Benchmark for Object
  Tracking in the Wild}.
\newblock In {\em ECCV}, pages 310--327, 2018.

\bibitem{GIoU}
Hamid Rezatofighi, Nathan Tsoi, JunYoung Gwak, Amir Sadeghian, Ian~D. Reid, and
  Silvio Savarese.
\newblock Generalized {Intersection Over Union}: {A} {Metric and a Loss for
  Bounding Box Regression}.
\newblock In {\em CVPR}, pages 658--666, 2019.

\bibitem{ImageNet}
Olga Russakovsky, Jia Deng, Hao Su, Jonathan Krause, Sanjeev Satheesh, Sean Ma,
  Zhiheng Huang, Andrej Karpathy, Aditya Khosla, and Michael Bernstein.
\newblock {ImageNet} {Large} {Scale Visual Recognition Challenge}.
\newblock {\em IJCV}, 115(3):211--252, 2015.

\bibitem{CSWinTT}
Zikai Song, Junqing Yu, Yi-Ping~Phoebe Chen, and Wei Yang.
\newblock Transformer tracking with cyclic shifting window attention.
\newblock In {\em CVPR}, pages 8791--8800, 2022.

\bibitem{googlenet}
Christian Szegedy, Wei Liu, Yangqing Jia, Pierre Sermanet, Scott Reed, Dragomir
  Anguelov, Dumitru Erhan, Vincent Vanhoucke, and Andrew Rabinovich.
\newblock Going deeper with convolutions.
\newblock In {\em CVPR}, pages 1--9, 2015.

\bibitem{SINT}
Ran Tao, Efstratios Gavves, and Arnold W.~M. Smeulders.
\newblock {Siamese Instance Search for Tracking}.
\newblock In {\em CVPR}, pages 1420--1429, 2016.

\bibitem{touvron2021training}
Hugo Touvron, Matthieu Cord, Matthijs Douze, Francisco Massa, Alexandre
  Sablayrolles, and Herv{\'e} J{\'e}gou.
\newblock Training data-efficient image transformers \& distillation through
  attention.
\newblock In {\em ICML}, pages 10347--10357, 2021.

\bibitem{2017Attention}
Ashish Vaswani, Noam Shazeer, Niki Parmar, Jakob Uszkoreit, Llion Jones,
  Aidan~N Gomez, Lukasz Kaiser, and Illia Polosukhin.
\newblock Attention is all you need.
\newblock In {\em NIPS}, pages 5998--6008, 2017.

\bibitem{wang2021transformer}
Ning Wang, Wengang Zhou, Jie Wang, and Houqiang Li.
\newblock Transformer {Meets Tracker: Exploiting Temporal Context for Robust
  Visual Tracking}.
\newblock In {\em CVPR}, pages 1571--1580, 2021.

\bibitem{SiamMask}
Qiang Wang, Li Zhang, Luca Bertinetto, Weiming Hu, and Philip H.~S. Torr.
\newblock {Fast Online Object Tracking and Segmentation}: {A} {Unifying
  Approach}.
\newblock In {\em CVPR}, pages 1328--1338, 2019.

\bibitem{wang2021pyramid}
Wenhai Wang, Enze Xie, Xiang Li, Deng-Ping Fan, Kaitao Song, Ding Liang, Tong
  Lu, Ping Luo, and Ling Shao.
\newblock Pyramid {Vision Transformer}: {A} {Versatile Backbone for Dense
  Prediction without Convolutions}.
\newblock In {\em ICCV}, pages 548--558, 2021.

\bibitem{wu2021cvt}
Haiping Wu, Bin Xiao, Noel Codella, Mengchen Liu, Xiyang Dai, Lu Yuan, and Lei
  Zhang.
\newblock {CvT}: {I}ntroducing {Convolutions} to {Vision} {Transformers}.
\newblock In {\em ICCV}, pages 22--31, 2021.

\bibitem{wu2022tinyvit}
Kan Wu, Jinnian Zhang, Houwen Peng, Mengchen Liu, Bin Xiao, Jianlong Fu, and Lu
  Yuan.
\newblock {TinyViT}: {Fast Pretraiobject ning Distillation for Small Vision
  Transformers}.
\newblock In {\em ECCV}, pages 68--85, 2022.

\bibitem{sbt}
Fei Xie, Chunyu Wang, Guangting Wang, Yue Cao, Wankou Yang, and Wenjun Zeng.
\newblock Correlation-aware deep tracking.
\newblock In {\em CVPR}, pages 8751--8760, 2022.

\bibitem{SiamFC++}
Yinda Xu, Zeyu Wang, Zuoxin Li, Ye Yuan, and Gang Yu.
\newblock {SiamFC++:} {T}owards {Robust and Accurate Visual Tracking with
  Target Estimation Guidelines}.
\newblock In {\em AAAI}, pages 12549--12556, 2020.

\bibitem{Stark}
Bin Yan, Houwen Peng, Jianlong Fu, Dong Wang, and Huchuan Lu.
\newblock {Learning Spatio-Temporal Transformer for Visual Tracking}.
\newblock In {\em ICCV}, pages 10428--10437, 2021.

\bibitem{yan2021lighttrack}
Bin Yan, Houwen Peng, Kan Wu, Dong Wang, Jianlong Fu, and Huchuan Lu.
\newblock {LightTrack: Finding Lightweight Neural Networks for Object Tracking
  via One-Shot Architecture Search}.
\newblock In {\em CVPR}, pages 15180--15189, 2021.

\bibitem{ostrack}
Botao Ye, Hong Chang, Bingpeng Ma, Shiguang Shan, and Xilin Chen.
\newblock {Joint Feature Learning and Relation Modeling for Tracking: A
  One-Stream Framework}.
\newblock In {\em ECCV}, pages 341--357, 2022.

\bibitem{yu2021high}
Bin Yu, Ming Tang, Linyu Zheng, Guibo Zhu, Jinqiao Wang, Hao Feng, Xuetao Feng,
  and Hanqing Lu.
\newblock High-{Performance Discriminative Tracking with Transformers}.
\newblock In {\em ICCV}, pages 9836--9845, 2021.

\bibitem{yuan2021tokens}
Li Yuan, Yunpeng Chen, Tao Wang, Weihao Yu, Yujun Shi, Zi-Hang Jiang,
  Francis~EH Tay, Jiashi Feng, and Shuicheng Yan.
\newblock {Tokens-to-Token ViT: Training Vision Transformers from Scratch on
  ImageNet}.
\newblock In {\em ICCV}, pages 538--547, 2021.

\bibitem{Deeper-wider-SiamRPN}
Zhipeng Zhang and Houwen Peng.
\newblock {Deeper and Wider Siamese Networks for Real-Time Visual Tracking}.
\newblock In {\em CVPR}, pages 4591--4600, 2019.

\end{thebibliography}
}
\clearpage

\end{document}